\documentclass[10pt,twocolumn,letterpaper]{article}

\usepackage[pagenumbers]{cvpr} % To force page numbers, e.g. for an arXiv version

\usepackage{graphicx}
\usepackage{amsmath}
\usepackage{amssymb}
\usepackage{booktabs}
\usepackage{bbm}

\long\def\ignorethis#1{}

\newcommand{\argmin}{\operatornamewithlimits{argmin}}

\usepackage{color}
\definecolor{gray}{rgb}{0.35,0.35,0.35}
\definecolor{green}{rgb}{0,1,0.2}
\definecolor{blue}{rgb}{0,0,1}
\definecolor{white}{rgb}{1,1,1}

\usepackage{xcolor}

\usepackage{epsfig}
\usepackage{epstopdf}
\usepackage{multirow}
\usepackage{array}
\usepackage{booktabs}

\newbox\jsavebox

\def\TTSw{TTS$_{\rm{weak}}$}
\def\TTSb{TTS$_{\rm{box}}$}
\def\TTSp{TTS$_{\rm{poly}}$}
\def\TTSs{TTS$_{\rm{synthetic}}$}

\usepackage[pagebackref,breaklinks,colorlinks]{hyperref}

\usepackage[capitalize]{cleveref}
\crefname{section}{Sec.}{Secs.}
\Crefname{section}{Section}{Sections}
\Crefname{table}{Table}{Tables}
\crefname{table}{Tab.}{Tabs.}

\def\cvprPaperID{2703} % *** Enter the CVPR Paper ID here
\def\confName{CVPR}
\def\confYear{2022}

\begin{document}

% \title{Reducing Supervision for End-to-End Text Spotting using a \\ Multi-Task Transformer}
\title{Towards Weakly-Supervised Text Spotting using a Multi-Task Transformer}
\vspace{-2.mm}
% \title{TextTranSpotter: a Multi-Task Transformer for End-to-End Text Spotting}
% \title{TextTranSpotter: a Multi-Task Transformer for Weakly-Supervised End-to-End Text Spotting}

\author{Yair Kittenplon \quad Inbal Lavi \quad Sharon \vspace{0.3mm}
Fogel \quad Yarin Bar\\ 
\vspace{2.7mm}
R. Manmatha \quad  Pietro Perona\\
AWS AI Labs\\
{\tt\small \{yairk, ilavi, shafog, yarinbar, manmatha, peronapp\}@amazon.com}
\\
}

\maketitle

%%%%%%%%% ABSTRACT
\begin{abstract}
Text spotting end-to-end methods have recently gained attention in the literature due to the benefits of jointly optimizing the text detection and recognition components. 
Existing methods usually have a distinct separation between the detection and recognition branches, requiring exact annotations for the two tasks.
We introduce TextTranSpotter (TTS), a transformer-based approach for text spotting and the first text spotting framework which may be trained with both fully- and weakly-supervised settings.
By learning a single latent representation per word detection, and using a novel loss function based on the Hungarian loss, our method alleviates the need for expensive localization annotations.
Trained with only text transcription annotations on real data, our weakly-supervised method achieves competitive performance with previous state-of-the-art fully-supervised methods.
When trained in a fully-supervised manner, TextTranSpotter shows state-of-the-art results on multiple benchmarks \footnote {Our code will be publicly available upon publication.}.
\end{abstract}

%%%%%%%%% BODY TEXT
\section{Introduction} \label{sec:intro}

Text spotting, \ie, detecting and reading text in images, 
is a key capability for machines to operate in the real world. Applications include vehicle navigation in buildings and cities, indexing of image collections and video, automated handling of packages, and prosthetics for blind and visually impaired people. This challenge was recognized early in the computer vision literature~\cite{findingtext, li2000automatic, yuille2004} and is currently undergoing a deep learning revival~\cite{TowardsET, jaderberg2014deep}, with most researchers focusing on two issues: architectures and data.
\begin{figure}[htbp]
\centering
\hspace{3.9mm}
\includegraphics[width=0.45\textwidth]{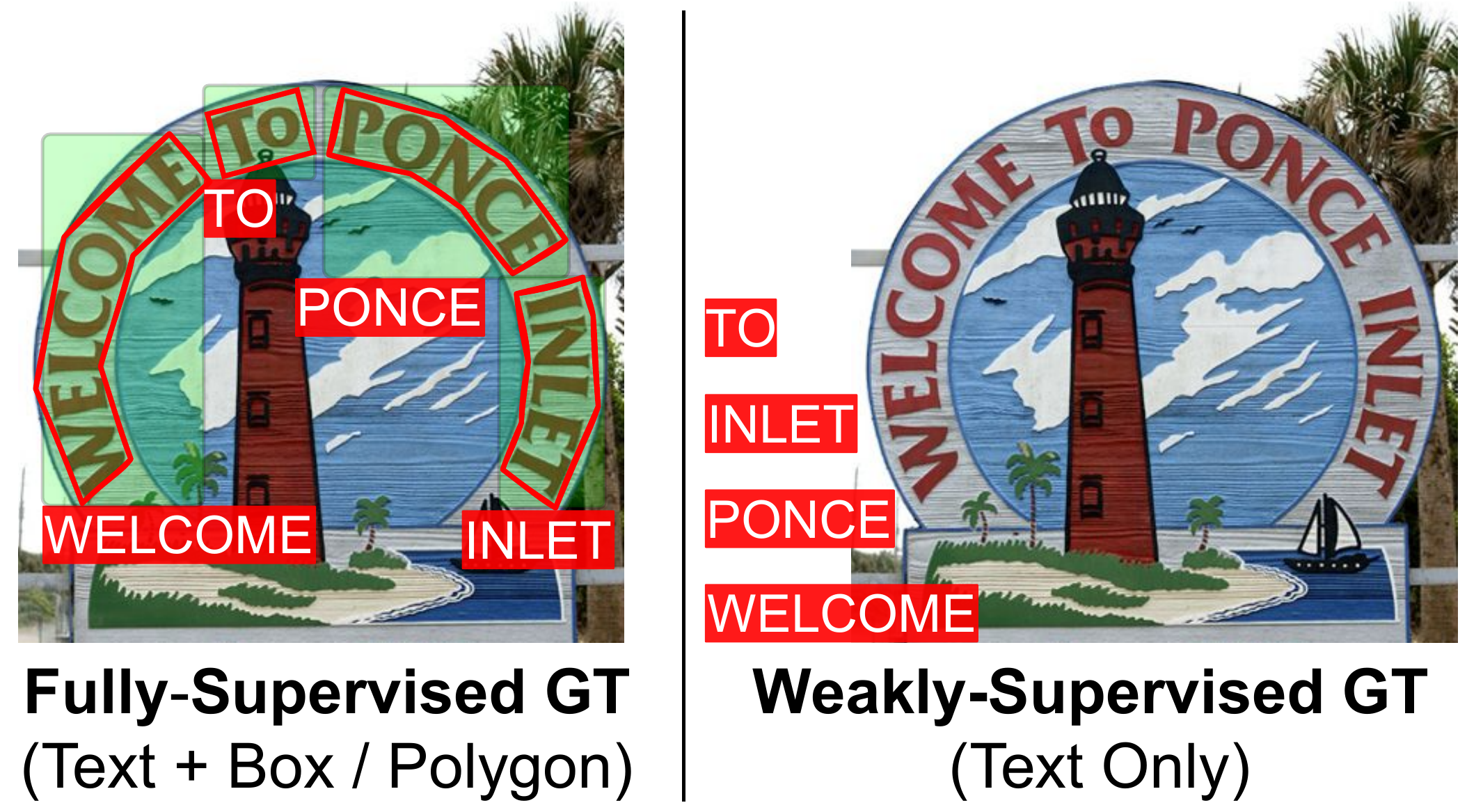}

\vspace{2.5mm}

\includegraphics[width=0.49\textwidth]{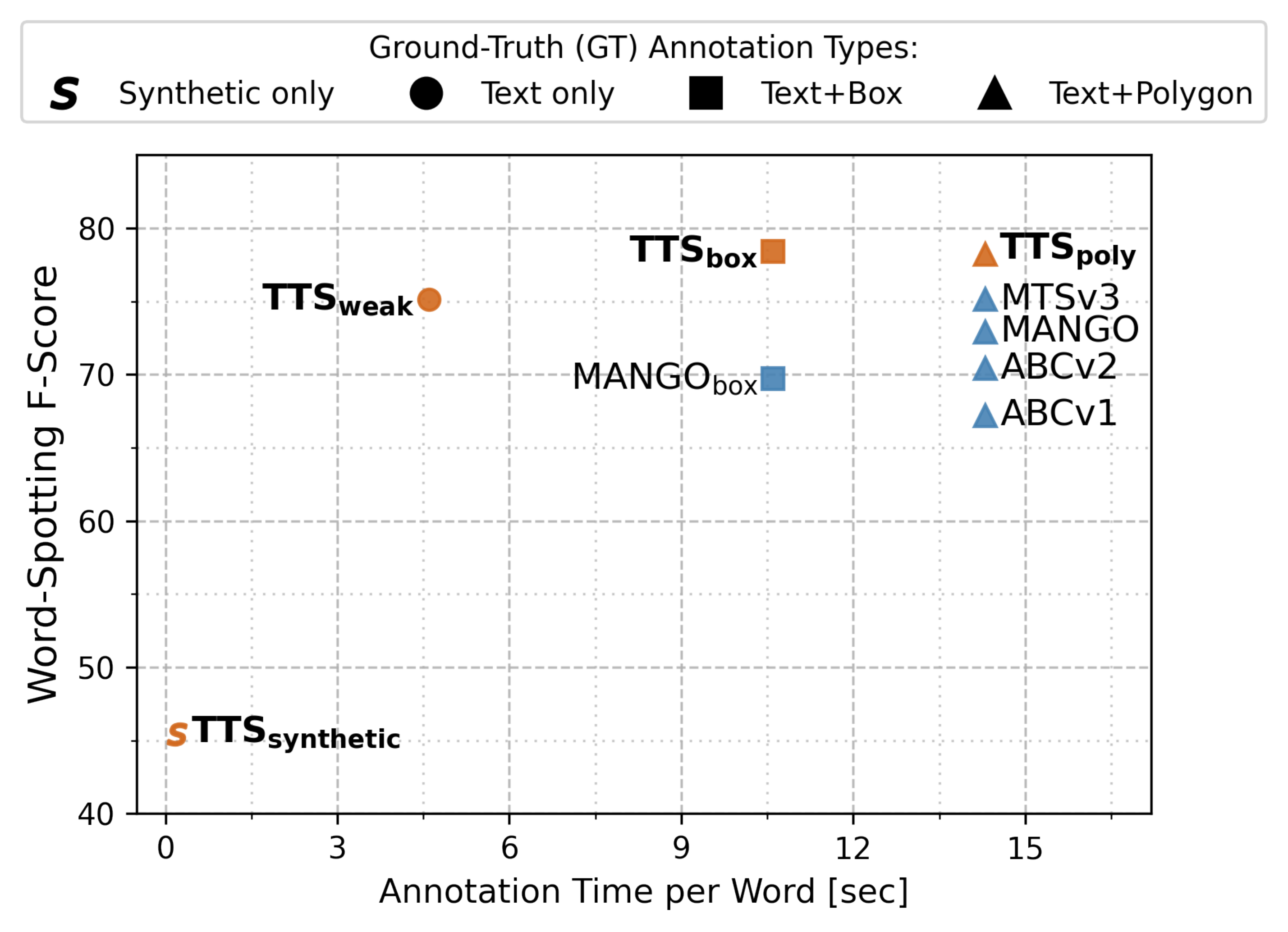}
\vspace{-1.mm}
\caption{\textbf{Weakly-supervised text spotting.} Top: Visualization of 'fully' (left) and 'weakly' (right) supervised ground-truth (GT) annotations. 
Bottom: 
Text spotting methods results on the Total-Text dataset (higher is better) vs. the time cost per word to annotate the datasets used for training (Sec. \ref{sec:annotations}, lower is better). Even when using weaker annotations only, our method surpasses state-of-the-art fully-supervised methods.
} 

\vspace{-3.5mm}

\label{fig:anno-time}
\end{figure}

Early systems \cite{jaderberg2016reading, busta2017deep} use separate architectures for text detection and recognition, without sharing any component.
More recent approaches take a leap forward towards a unified end-to-end architecture by sharing a convolutional feature backbone \cite{survey, TowardsET} and employing a feature cropping mechanism to extract the relevant area of interest for the recognition head. Such architectures are still not ideal, since the recognition head is usually trained using the detection ground-truth and thus it is not optimized for the predictions of the detection head. 
Furthermore, the detection head is trained as a standard object detection model, without regard to the additional supervision given by the text transcription or to the downstream recognition task. Other than mutually optimizing the backbone, the tasks are separate, requiring both transcription annotations for the recognition head, and polygons or bounding box annotations for the detection head.
Recently, more sophisticated methods forgo the two-stage approach by directly localizing and classifying the characters in the text~\cite{crafts, mango}, which further requires character-level annotations.

The datasets in the field of text spotting consist of synthetic and real data. Real data annotation is an expensive task, however relying solely on synthetic data leads to poor results. Most of the annotation time is dedicated to the detection ground-truth, while the transcription annotation alone requires less than half of the time, as discussed in Sec. \ref{sec:annotations}.
State-of-the-art-methods explicitly segment the text area, allowing the recognizer to cope with rotated, curved, or densely located text and ignore background noise~\cite{survey, spotterV3}. A disadvantage of such methods is that they require expensive polygonal annotations \cite{icdar2015, totaltext}.

In this work, we suggest a new text spotting approach, TextTranSpotter (TTS), which can forgo the expensive spatial annotations and use only transcript annotations for real data. This setting is \emph{weakly supervised}, in the sense that only partial information about the text in the image is used for training. At inference time, the model outputs both the detection and the transcription of the text in the image. The weakly supervised setting has many use-cases, especially in situations where annotation resources are limited or there is an existing dataset with only text transcription annotations \cite{books}.
Furthermore, TTS can be trained in both a fully- or a weakly-supervised manner, thus allowing a trade-off between model performance and annotation cost (Fig. \ref{fig:anno-time}).

\begin{figure*}
\begin{center}
\includegraphics[width=0.995\linewidth]{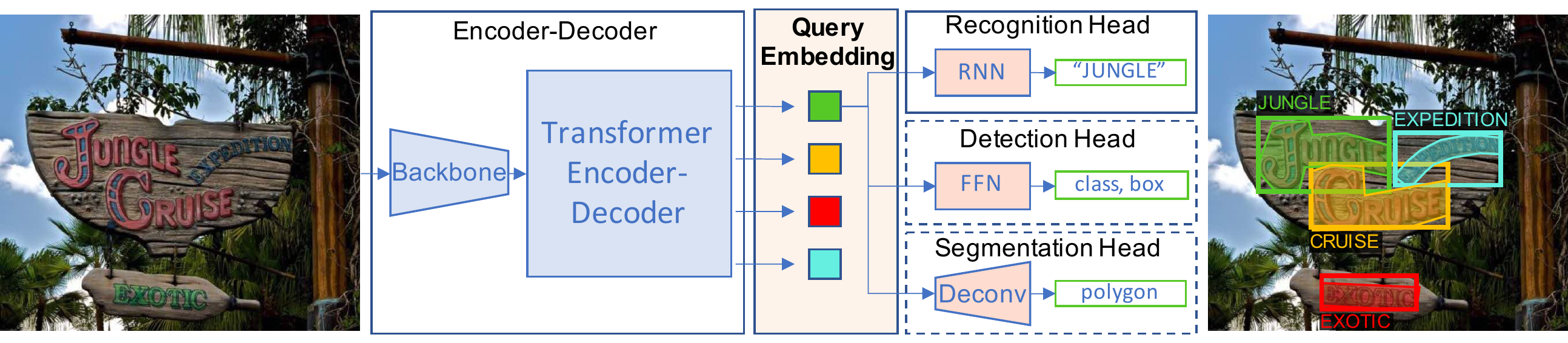}
\end{center}
\vspace{-5.5mm} 
   \caption{\textbf{TextTranSpotter.} An overview of our end-to-end architecture. Unlike previous approaches which share only the backbone, in TextTranSpotter the transformer encoder-decoder computes a joint query embedding for each detection (colored square). This embedding is shared for both recognition, detection and segmentation heads, which consist of a recurrent neural network (RNN), a linear feed-forward network (FFN) and a deconvolutional decoder (Deconv), respectively. Our weakly-supervised setting works by training only the recognition and classification heads on the real data, using the detection head at inference time for box prediction.
   An illustration comparing our architecture with previous text-spotting approaches can be found in the supplementary.}
\label{fig:arch}
\vspace{-2.5mm} 
\end{figure*}

To allow the weakly-supervised setting, we depart from existing text spotting methods which treat text detection and recognition as related but independent tasks. Our approach includes a novel architecture and loss function which better entangle the two tasks, taking a step further towards a unified end-to-end system.
TextTranSpotter takes advantage of recent developments in transformers \cite{vaswani2017attention, dosovitskiy2020image, detr} to create a multitask network (see Fig. \ref{fig:arch}), learning a single object query embedding for both detection and recognition heads. The task heads are very simple and lean; the detection head is a linear feed-forward network and the recognition head is a Recurrent Neural Network (RNN) \cite{crnn}. This indicates that the majority of the computation is performed in the shared transformer, unlike most approaches which use more intricate recognition and detection networks (see supplementary for comparison of methods).
The input to the recognition head is the transformer output, which allows it to learn the relevant areas of interest for the given query instead of being given this area explicitly as input. Therefore, it does not require accurate segmentation of the text to perform even in challenging scenarios, such as rotated text, arbitrary-shaped text, or text with overlapping bounding boxes (Sec. \ref{sec:Robustness}).
If the segmentation output is desired, a mask head can be added similarly to the detection and recognition head using a simple deconvolutional decoder.

Our weakly-supervised training scheme is obtained by introducing a new loss function based on the Hungarian matching loss~\cite{detr} that simultaneously optimizes the detection and recognition tasks. 
The Hungarian loss, which has shown promise in the field of object detection \cite{detr, defdetr, updetr, ming2021optimization}, is meaningful in our setting, where the matching explicitly uses the text content for the detection optimization. Our Hungarian loss, which we call Text Hungarian Loss, replaces the detection cost with a recognition cost in the matching criteria. The shared embedding that is optimized in this manner allows for a significant benefit compared to training on synthetic data only, without using any spatial information about the real data. Our weakly-supervised model reaches results comparable to existing fully-supervised methods.

Our main contributions are:

\begin{enumerate}
\item A weakly-supervised training scheme using only the text annotations without any spatial ground-truth for real data, utilizing a novel text-based Hungarian matching loss.

\item The first multi-task transformer-based approach for text spotting, in which a single representation is being learned per word for both detection and recognition predictions.

\item Extensive quantitative benchmarks showing our fully-supervised method achieves state-of-the-art results on common text spotting benchmarks, and our weakly-supervised method achieves results competitive with previous fully-supervised methods.

\item The first text spotting framework to offer both a fully-supervised training scheme and a weakly-supervised one for the same architecture, presenting a trade-off between model accuracy and annotation cost.

\end{enumerate}
\section{Related Work} \label{sec:related}
\noindent\textbf{Text Spotting.}
Li \etal \cite{TowardsET} may be the first to integrate deep detection and recognition modules into a unified end-to-end system, by using a shared backbone encoder and RoIPooling \cite{faster_rcnn} to feed the detected features to the recognition head. Liu \etal \cite{FOTS} suggest using RoIRotate to enable feature extraction from rotated rectangle detection results. Liao \etal \cite{spotterV1} introduce Mask TextSpotter, which takes advantage of character-level annotations to detect and recognize characters and instance masks, in order to handle arbitrary-shaped scene text.  Xing \etal \cite{charnet} detect and recognize individual characters, using the text instance detection results to group them.
Liu \etal \cite{abc} fit parameterized Bezier curves to the text contour, and design a BezierAlign layer for curved text feature extraction.
Qin \etal \cite{qin19towards} propose RoiMask, focusing on the arbitrary-shaped text region. Feng \etal \cite{dragon}
suggest using RoISlide, a sampling method which fuses features from the predicted segments of the text, allowing robustness to long arbitrary-shaped text. Liao \etal \cite{spotterV3} improve Mask TextSpotter \cite{spotterV1} by adding a Segmentation Proposal Network (SPN) to generate proposals represented by accurate polygons. Qiao \etal \cite{mango} remove the RoI operations and design a position-aware attention module to coarsely localize the text sequences. However, character-level and polygon annotations are required. Baek \etal \cite{crafts} also learn character-level masks, which are fed into an attention-based recognizer.

We adopt the idea of an  end-to-end system, and further suggest a unified encoding-decoding mechanism, based on a multi-task transformer.
Learning a mutual feature embedding per query frees us from the need to design a hand-crafted feature pooling operation.
Furthermore, the multi-task nature of our method alleviates the need for exact annotations such as polygons or character-level annotations.

\noindent\textbf{Weakly Supervised Approaches.} Zhao \etal \cite{weak_detection} suggest a weakly-supervised approach for arbitrary text detection, by using an Expectation-Maximization based method, and provide an extensive study of the annotation time under different supervision levels. Janouskova \etal \cite{books} generate a large dataset for text recognition out of weakly-annotated existing data by using a pre-trained localization module as its annotator. In order to create pseudo ground truth labels, they use Levenshtein distance to match predicted transcriptions to a weakly annotated ground-truth set.
Bartz \etal \cite{see} suggest training an actual end-to-end text spotting system in a weakly supervised manner, by using a fixed resolution grid as a differentiable localization pooling mechanism. 
Qiao \etal \cite{mango} take a step in the direction of weakly-supervised text spotting by training with bounding boxes instead of polygons, but this results in a significant reduction in performance.

Motivated by the study of Zhao \etal \cite{weak_detection}, showing the high cost of polygonal or segmentation masks annotations, we suggest an end-to-end recognition method in which bounding boxes are sufficient for the task. 
Moreover, we introduce a weakly-supervised framework, in which text transcriptions are the only real-data annotations needed for training, and provide a study of annotation times for both detection and text transcription.

\noindent\textbf{Hungarian Matching.}
Throughout the past decade, learning based approaches for object detection \cite{faster_rcnn, yolo, ssd, fcos} have been used to learn engineered dense predictions, and filter near-duplicate predictions using hand-crafted rules.
Recently, Carion \etal \cite{detr} presented a new object detection method, DETR, that formulates the problem as a direct set prediction problem. It uses a bipartite matching loss based on the Hungarian algorithm \cite{hungarian} to perform a one-to-one matching between ground-truth and predicted detections, unlike dense approaches in which the matching is one-to-many. This sparse detection paradigm has become popular in the object detection literature \cite{sparse_rcnn, defdetr, sun2020rethinking} and has advanced the field. Zhu \etal \cite{defdetr} mitigate some of the issues in DETR, namely the slow convergence and low performance on small objects, by incorporating deformable attention and a multi-scale architecture.

Following this line of research, we find the sparse detection approach suitable for multi-task loss formulation, where a given object query can be optimized for additional tasks besides detection.
We use a Hungarian matching based loss, by adding a recognition cost term to the matching criteria.
\section{Method} \label{sec:method}
We suggest an end-to-end text spotting approach, named TextTranSpotter (TTS). A description of its architecture is presented in Sec.~\ref{sec:arch}, a novel variation of the Hungarian matching loss for text spotting is described in Sec.~\ref{sec:hungarian}, and an adaptation of this method into a weakly-supervised setting is described in Sec.~\ref{sec:weakly}.
\subsection{Architecture}
\label{sec:arch}
TTS consists of a transformer-based Encoder-Decoder, followed by parallel detection, recognition, and segmentation heads, as illustrated in Fig.~\ref{fig:arch}.

\noindent\textbf{Joint Query Embedding.} Our Encoding-Decoding module is shared between the detection and recognition branches. Following Carion \etal \cite{detr}, our architecture uses a predetermined number of learned positional embeddings as input to the decoder, called object queries. The decoder learns a latent representation per object query, $q_{emb}\in {\mathbb R}^{d_{emb}}$, which is used as an input to all the task-specific heads in our model. Both detection and recognition heads are designed in a light-weight manner, meaning that the majority of the computation is done by the transformer which is optimized jointly for both tasks. This setting improves the embedding for the detection task not only through the detection loss but through the recognition loss optimization, as we show in the ablation study in Sec. \ref{sec:ablation}. If a polygon output is desired, the optimized query embedding can be used as an input to a segmentation head, as described below.

The network is based on the Deformable-DETR architecture for object detection \cite{defdetr}. It consists of a conventional CNN backbone that generates a multi-scale feature map, followed by a deformable transformer encoder-decoder, in which the offsets of the attention heads are learned in addition to the attention maps themselves. The dynamic structure of this attention mechanism enables the recognition of rotated, curved, and even upside-down text, without any special treatment as described in Sec. \ref{sec:Robustness}. In fact, our network is able to achieve this even though it is trained using only axis-aligned box annotations, which are significantly less costly than polygon annotations.

\noindent\textbf{Detection Head.} We follow recent object detection methods \cite{detr, defdetr}, and use a 3-layer feed-forward network (FFN) to regress the normalized parameters of the query word box w.r.t. the input image, and a linear projection layer to predict the query score, \ie, classify whether or not a query contains a word.

\noindent\textbf{Recognition Head.} 
To the best of our knowledge, all previous recognition models, including ones used in text spotting approaches, use a spatial signal as input to the recognition head (\eg, an image, or a cropped output of the backbone). In our approach, only the one-dimensional joint query embedding, computed by the transformer encoder-decoder, is used. To extract the text transcription we use a sequential LSTM-based decoder, with a one-to-many mapping where the input is the joint query embedding $q_{\rm{emb}}$ and the output for each time-step $k$ is the character probabilities $t^k\in {\mathbb R}^{l}$ where $l$ is the length of the alphabet.

\noindent\textbf{Segmentation Head.} 
TTS is trained in its fully-supervised setting using the text bounding boxes and recognition transcriptions, without any polygon annotations.
However, if a polygon output is desired, a segmentation head can be trained separately based on the frozen TTS model weights. Given the pre-trained query embedding, a light-weight segmentation head, built with 4 linear layers and 3 deconvolution layers, may be used to extract a binary mask, describing the text in the detected bounding box. 
A polygonal output is then computed from the binary mask.
\begin{figure}
\begin{center}
\includegraphics[width=0.9\linewidth]{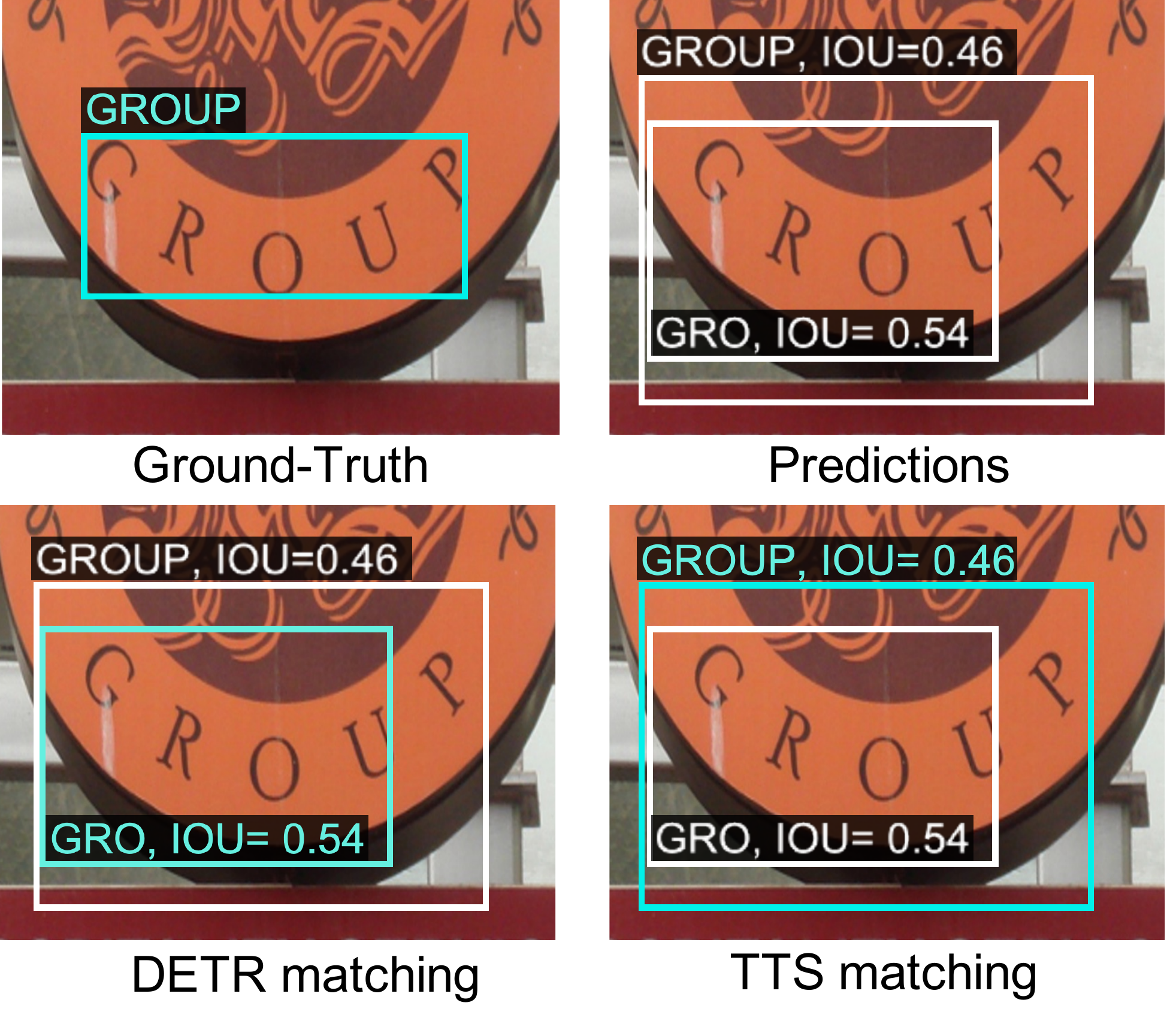}
\end{center}
\vspace{-5.mm}

\caption{\textbf{Matching operation.}
Top: GT and model predictions during training.
Bottom: Matching operation using the original DETR criteria \cite{detr} (left), and using our suggested criteria (right).
The prediction matched with the GT is marked in blue.
It can be seen that TTS matches the prediction with the best predicted transcription, even though its box IOU score is lower.}

\vspace{-3.mm} 

\label{fig:illustration}
\end{figure}

\subsection{Text Hungarian Loss}
\label{sec:hungarian}
Inspired by recent object detection approaches \cite{detr, sparse_rcnn, defdetr}, we adopt the bipartite matching loss approach, using the Hungarian algorithm \cite{hungarian} to find a one-to-one matching $\hat{\sigma}$, between the ground-truth and predicted detections:
\begin{equation}
\hat{\sigma} = \argmin_{\sigma \in \theta_N} \sum_{i=1}^N {C(y_i, \hat{y}_{\sigma (i)})},  
\end{equation}
where $C$ is the criteria used to perform the matching, $y$ is the ground truth set, $\hat{y}$ is the predicted set, $N$ is the number of predictions, or object queries, and $\theta_N$ is the set of possible matches.
\noindent The Hungarian loss function is formulated based on the matching  $\hat{\sigma}$:
\begin{equation}
L_{\rm{Hungarian}}(y, \hat{y}) = \sum_{i=1}^N L(y_i, \hat{y}_{\hat{\sigma}(i)}).
\end{equation}

To better leverage the transcription annotations, we take into account not only the detection and classification criteria as in \cite{detr}, but also add a recognition-based criteria $C_{rec}$, and loss $L_{rec}$, into the matching cost $C$ and loss $L$, introducing a novel Text Hungarian Loss.

The fully-supervised matching criteria is:
\begin{multline}
C(y, \hat{y}_{\sigma (i)}) = -\alpha_c \hat{p}_{\sigma(i)}(c_i)+\mathbbm{1}_{\{c_i \neq \emptyset\}} \alpha_{\rm{box}} C_{\rm{box}}(b_i,\hat{b}_{\sigma(i)}) \\ + \mathbbm{1}_{\{c_i \neq \emptyset\}} \alpha_{\rm{rec}} C_{\rm{rec}}(t_i,\hat{t}_{\sigma(i)}),
\end{multline}

\noindent where $c_i$, $b_i$ and $t_i$ are the ground truth class, bounding box and transcription respectively, $\hat{p}_{\sigma(i)}(c_i)$ is the predicted probability for class $c_i$, and $\alpha_c$, $\alpha_{box}$ and $\alpha_{rec}$ are the weights for the classification, bounding box, and transcription criteria. The fully-supervised loss term is:
\begin{multline}
L(y_i, \hat{y}_{\hat{\sigma}(i)}) = -\beta_c log \hat{p}_{\hat{\sigma}(i)}(c_i) \\
+ \mathbbm{1}_{\{c_i \neq \emptyset\}} \beta_{\rm{box}} L_{\rm{box}}(b_i,\hat{b}_{\hat{\sigma}(i)}) 
+\mathbbm{1}_{\{c_i \neq \emptyset\}} \beta_{\rm{rec}} L_{\rm{rec}}(t_i,\hat{t}_{\hat{\sigma}(i)}).
\end{multline}

\begin{figure}[t!]
\centering
\includegraphics[width=0.45\textwidth]{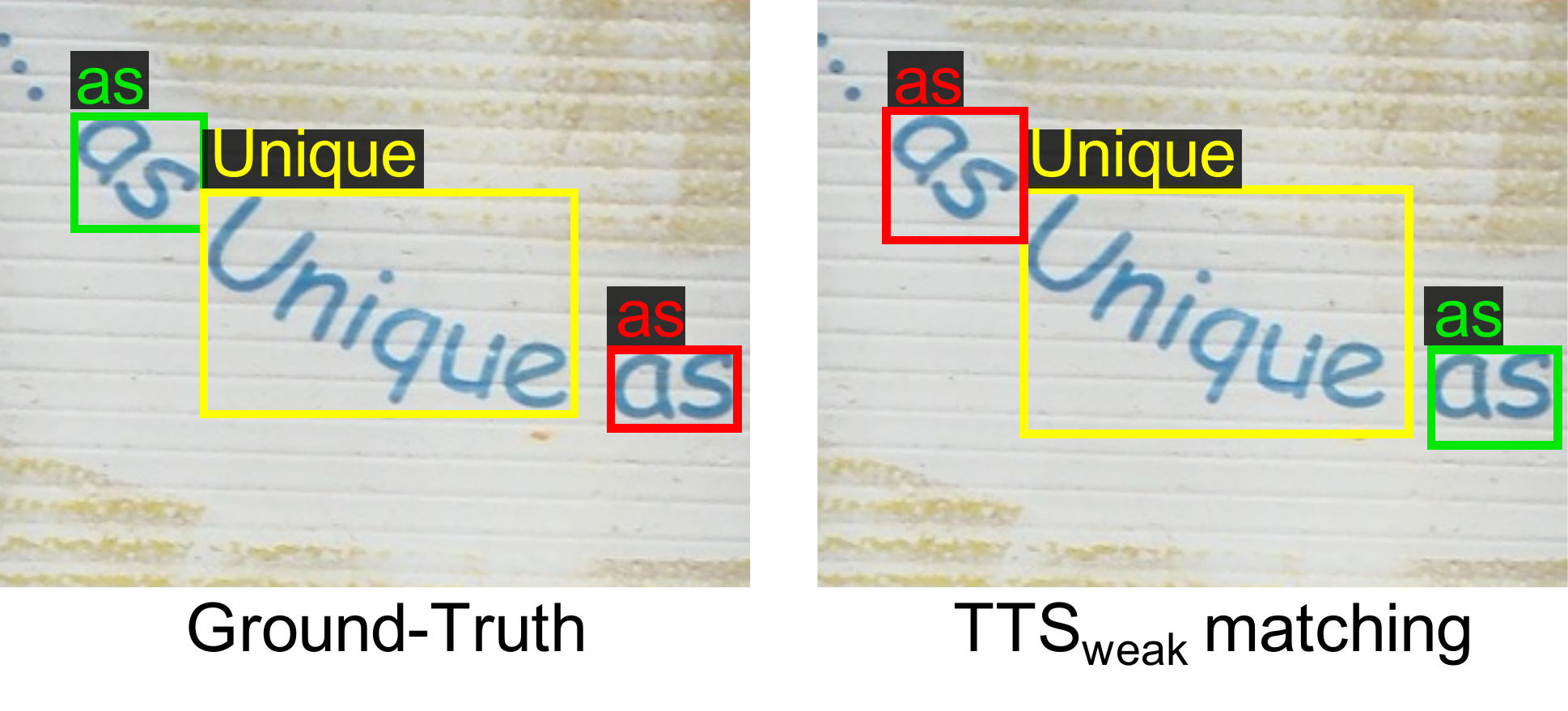}
\vspace{-2.mm}
\caption{\textbf{Weakly-supervised matching swap.} 
GT instances are shown on the left. Predictions during the training of \TTSw~ are shown on the right, where each prediction is matched with the GT of the same color. Although the matching of the two occurrences of the word ``as'' are swapped, the weakly-supervised loss remains the same (Eq. \ref{eq:loss_weakly}), hence the matching is correct.
}
\vspace{-1.5mm} 
\label{fig:swap}
\end{figure}

Where $L_{\rm{box}}$ is the bounding box loss, defined as in DETR \cite{detr}, and $\beta_c$, $\beta_{box}$ and $\beta_{rec}$ are the weights for the classification, bounding box, and transcription losses.

We use a cross entropy loss for both the recognition criteria and loss terms:
\begin{equation}
    C_{\rm{rec}}(t_i,\hat{t}_{\sigma(i)}) = L_{\rm{rec}}(t_i,\hat{t}_{\sigma(i)}) = \sum_j -log \hat{p}_{\sigma(i)}(t_i^j)
\end{equation}
where $j$ is the character index in the word $t_i$.

Fig. \ref{fig:illustration} shows examples of matching between the ground-truth and the model's predictions using different criteria.
Using only the detection and classification scores for the matching, as in DETR \cite{detr}, may lead the model to match a box query with a higher intersection-over-union (IOU) but worse recognition results.

We experiment with the new loss and various settings for the new matching term, as described in Sec. \ref{sec:experiments}, and show that the addition of the recognition term contributes to better recognition performance in the end-to-end results.

\subsection{Weakly Supervised Text Spotting}
\label{sec:weakly}
Our Text Hungarian Loss finds a matching between ground-truth and predictions based not only on the detected box but also on the recognition output. This opens up the possibility to match the ground-truth and predicted words based only on the recognition and classification criteria. As a result, the model can be optimized using only the transcription annotations, \ie, a list of words that appear in the image, without any spatial annotations. At inference time the model still outputs bounding boxes for the predicted words, similarly to the fully-supervised models. We train the models in this setting with fully-supervised synthetic data, and weakly-supervised real (non-synthetic) data.

The criteria used in the weakly supervised training is therefore:
\begin{multline}
C_{\rm{weak}}(y, \hat{y}_{\sigma (i)}) = -\alpha_c \hat{p}_{\sigma(i)}(c_i) \\
+ \mathbbm{1}_{\{c_i \neq \emptyset\}} \alpha_{\rm{rec}} C_{\rm{rec}}(t_i,\hat{t}_{\sigma(i)})
\end{multline}

\noindent and the loss term is:
\begin{multline}
L_{\rm{weak}}(y_i, \hat{y}_{\hat{\sigma}(i)}) =\\ -\beta_{c}log \hat{p}_{\hat{\sigma}(i)}(c_i)+ \mathbbm{1}_{\{c_i \neq \emptyset\}} \beta_{\rm{rec}} L_{\rm{rec}}(t_i,\hat{t}_{\hat{\sigma}(i)}).
\label{eq:loss_weakly}
\end{multline}

Note that in this setting, if there are multiple words with the same transcription, it is possible that there is more than a single correct match. An example of this case is shown in Fig. \ref{fig:swap}, where the word ``as'' repeats twice in the image causing the queries to be mismatched. Since the training is performed only on the recognition head and not the bounding-box regression, the supervision for each of the transcriptions remains the same and does not affect the training process.

\section{Experiments} \label{sec:experiments}
We evaluate TextTranSpotter on common benchmarks using both the fully- and weakly-supervised settings. We further test the model performance on rotated and curved text, and conduct ablation studies regarding 
its architecture and matching criteria.

\begin{table*}
\centering
\begin{tabular}{@{}l|cccccc|cccc@{}}
\toprule
\multirow{3}{*}{Method} & \multicolumn{6}{c|}{ICDAR 2015}                                     & \multicolumn{4}{c}{Total-Text}                                     \\ \cmidrule(l){2-11} 
                        & \multicolumn{3}{c}{Word Spotting} & \multicolumn{3}{c|}{End-to-End} & \multicolumn{2}{c}{Word Spotting} & \multicolumn{2}{c}{End-to-End} \\ \cmidrule(l){2-11} 
                        & S         & W         & G         & S         & W        & G        & None             & Full           & None           & Full          \\ \midrule
MTS-V1 \cite{spotterV1}                  & 79.3      & 74.5      & 64.2      & 79.3      & 73.0     & 62.4     & -                & -              & 52.9           & 71.8          \\
MTS-V2 \cite{spotterv2}                 & 82.4      & 78.1      & 73.6      & 83.0      & 77.7     & 73.5     & -                & -              & 65.3           & 77.4          \\
TextDragon \cite{dragon}             & \textbf{86.2}      & \textbf{81.6}      & 68.0      & 82.5      & 78.3     & 65.2     & -                & -              & 48.8           & 74.8          \\
ABCNet-V1 \cite{abc}                 & -         & -         & -         & -         & -        & -        & 67.2             & 76.4           & 63.7           & 76.6          \\
MTS-V3 \cite{spotterV3}                 & 83.1      & 79.1      & \underline{75.1}      & 83.3      & 78.1     & 74.2     & \underline{75.1}             & 81.8           & 71.2           & 78.4          \\
ABCNet-V2 \cite{abc2}              & -         & -         & -         & 82.7      & 78.5     & 73.0     & 70.4             & 78.1           & -              & -             \\
CRAFTS \cite{crafts}                 & -         & -         & -         & 83.1      & \textbf{82.1}     & \underline{74.9}     & -                & -              &\textbf{ 78.7}           & -             \\
MANGO* \cite{mango}                  & \underline{85.2}      & 81.1      & 74.6      & \textbf{85.4}      & 80.1     & 73.9     & 72.9             & \underline{83.6}           & 68.9           & \underline{78.9}          \\
\textbf{\TTSp}                   & 85.0       & \underline{81.5}       & \textbf{77.3}       & \underline{85.2}       & \underline{81.7}      & \textbf{77.4}      & \textbf{78.2}               & \textbf{86.3}           & \underline{75.6}           & \textbf{84.4}          \\ \bottomrule
\end{tabular}
\vspace{-2.mm}
\caption{\textbf{Evaluation results on ICDAR 2015 and Total-Text datasets.} Word spotting and end-to-end f-score using strong (S), weak (W), generic (G), none and full lexicons. * MANGO \cite{mango} evaluated with IOU 0.1. Our method shows the best results using generic lexicons.}
\label{table:poly}
\vspace{-1.5mm}
\end{table*}
\begin{table*}
\centering
\begin{tabular}{@{}l|ccc|cccccc|cccc@{}}
\toprule
\multirow{3}{*}{Method} & \multicolumn{3}{c|}{\multirow{2}{*}{Annotations (real)}} & \multicolumn{6}{c|}{ICDAR 2015}                                     & \multicolumn{4}{c}{Total-Text}                                     \\ \cmidrule(l){5-14} 
                        & \multicolumn{3}{c|}{}                             & \multicolumn{3}{c}{Word Spotting} & \multicolumn{3}{c|}{End-to-End} & \multicolumn{2}{c}{Word Spotting} & \multicolumn{2}{c}{End-to-End} \\ \cmidrule(l){2-14} 
                        & Text            & Box            & Poly           & S         & W         & G         & S         & W        & G        & None             & Full           & None           & Full          \\ \midrule
\TTSs                    &                 &                &                & 53.1      & 46.9      & 42.9      & 53.2      & 47.0     & 43.0     & 45.4             & 60.9           & 46.3           & 58.8          \\  
\textbf{\TTSw}                    & $\pmb{\checkmark}$           &                &                & 78.6      & 75.1      & 70.2      & 78.7      & 75.2     & 70.1     & 75.1             & 83.5           & 71.5           & 80.1          \\
\textbf{\TTSb}                    & \checkmark               & \checkmark              &                & \textbf{84.9}      & \textbf{81.3}      & \textbf{77.1}      & \textbf{85.0}      & \textbf{81.5}     & \textbf{77.1}     & \textbf{78.4}             & \textbf{86.6}           & \textbf{75.8}           & \textbf{84.5}   
\\ \midrule 
$\textnormal{MANGO}_{box}$ \cite{mango}             & \checkmark               & \checkmark              &                & -         & -         & -         & -         & -        & -        & 69.7             & 80.6           & -              & -  \\
MTS-V3 $\dagger$ \cite{spotterV3}                 & \checkmark               & \checkmark              & \checkmark              & 82.7      & 78.5      & 74.7      & 82.5      & 77.4     & 73.5     & 74.8             & 81.2           & 70.5           & 77.7          \\\bottomrule
\end{tabular}
\vspace{-2.mm}
   \caption{\textbf{Results under limited training data annotations.} Word spotting and end-to-end f-score using strong (S), weak (W), generic (G), none and full lexicons. ``Text'', ``Box'' and ``Poly'' denotes the model trained on real data using annotations of text, bounding boxes and polygons, respectively. Axis-aligned evaluation was used. Results are improved dramatically when training on real data with text-only annotations (\TTSw) compared to training only with fully-supervised synthetic data (\TTSs). Using box annotations (\TTSb) improves results even further. $\dagger$ We show the results of MTS-V3\cite{spotterV3} using axis-aligned evaluation as a reference point.}
\label{table:box}
\vspace{-3.5mm}
\end{table*}
\begin{figure*}
\centering
\includegraphics[width=0.98\textwidth]{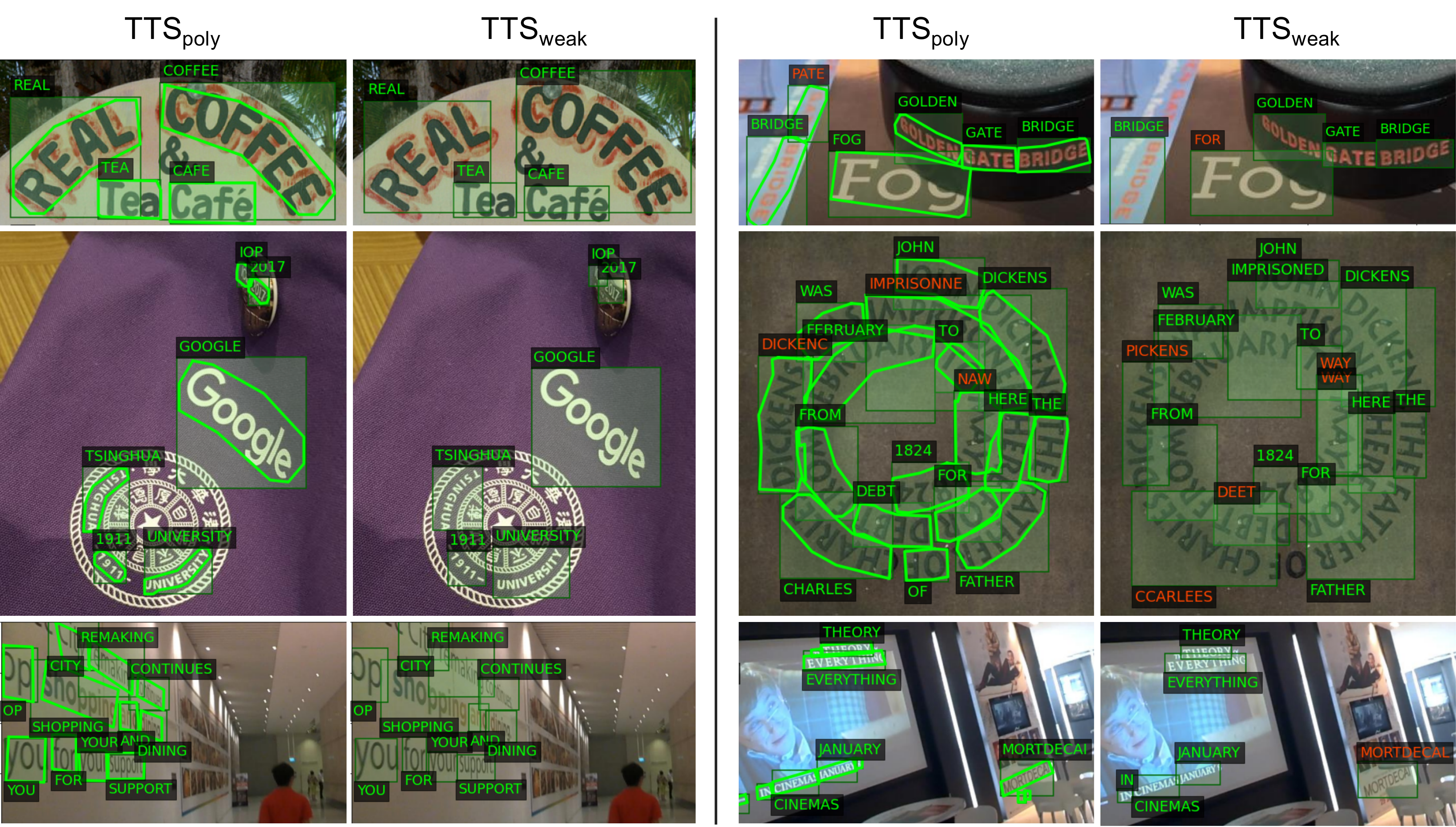}
\vspace{-2.mm}
\caption{\textbf{Qualitative results.} Prediction examples of our weakly-supervised (\TTSw) and fully-supervised (\TTSp) models, on Total-Text and ICDAR 2015 samples. TTS can handle rotated, curved, and even upside-down text instances, effectively distinguishing between overlapping boxes by extracting only the relevant text from the bounding box.
\TTSw~ has lower performance than \TTSp, which is expected given the reduction in supervision, however it manages to output high quality results. Fail cases are presented on the right.}
\label{fig:results}
\vspace{-2.mm}
\end{figure*}
\subsection{Implementation Details}
Following Liao \etal \cite{spotterV3}, the model is first trained on SynthText \cite{synthtext}, a large synthetic dataset with over 850k images, designed for both detection and recognition of text in images to obtain \TTSs. We then train our model on a mix of SynthText together with real datasets; Total-Text \cite{totaltext}, about 1k images including mainly curved text in various orientations and shapes, ICDAR 2015 \cite{icdar2015}, 1k images containing mostly small text instances, ICDAR 2013 \cite{ICDAR13}, 229 training images with mostly near-horizontal text, COCOText \cite{cocotext}, 43k train images taken from the MS-COCO dataset \cite{mscoco}, and SCUT \cite{scut}, 1k training images containing varied text.
Both our weakly- (\TTSw) and fully-supervised (\TTSb) models are obtained using this setup, where for \TTSw~ we use fully-annotated synthetic data, and weakly-annotated real data.
To produce a polygonal output, we freeze \TTSb~ weights, and train only a segmentation head, using the same mix of real datasets, and a subset of the SynthText dataet, with polygonal annotations. We call this model \TTSp.

We use Total-Text and ICDAR 2015 test data to evaluate both our fully-supervised and weakly-supervised models.
To test our method's robustness to rotations, we use the Rotated ICDAR 2013 dataset similarly to Liao \etal \cite{spotterV3}.

\subsection{Comparison to Previous Methods}
The evaluation results of TextTranSpotter, compared to previous approaches on Total-Text and ICDAR 2015, are shown in Table \ref{table:poly}. Evaluation was done using the standard polygonal evaluation protocol with IOU threshold of 0.5.

Both word spotting and end-to-end results are presented. For ICDAR 2015 we use ``strong'', ``weak'' and ``generic'' dictionaries, and for Total-Text, we show the results without a lexicon and using a ``full'' lexicon. 
On the Total-Text dataset, our method outperforms previous approaches with and without using a lexicon in the word spotting setting and using a ``full'' lexicon in the end-to-end setting.
On ICDAR 2015, our method shows the best results using ``generic'' lexicon, the most common and challenging use-case.

\begin{table}[]
\centering
\begin{tabular}{@{}l|cc|cc@{}}
\toprule
\multirow{2}{*}{Method} & \multicolumn{2}{c|}{$45^{\circ}$} & \multicolumn{2}{c}{$60^{\circ}$} \\ \cmidrule(l){2-5} 
                        & Det.       & E2E       & Det.      & E2E       \\ \midrule
CharNet  R-50 \cite{charnet}          & 57.2            & 33.9      & 58.8           & 9.3       \\
MTS-V2 \cite{spotterV1}                 & 62.2            & 54.2      & 65.5           & 56.6      \\
MTS-V3 \cite{spotterV3}                 & 84.2            & 76.1      & 84.7           & 76.6      \\
\textbf{\TTSp}                 & \textbf{88.8}            & \textbf{80.4 }     & \textbf{87.6}           & \textbf{80.1}      \\ \midrule
MTS-V3$\dagger$ \cite{spotterV3}                 & 82.9             & 75.4       & 81.2            & 75.3       \\
\textbf{\TTSb$\dagger$}                   & \textbf{89.9}            & \textbf{80.1}      & \textbf{89.7}           & \textbf{81.0}      \\ \bottomrule
\end{tabular}
\vspace{-1.5mm} 
   \caption{\textbf{Results on Rotated ICDAR 2013 dataset.} F-measure of detection (Det.) and end-to-end (E2E) recognition, under different rotation angles. $\dagger$ means that axis-aligned evaluation was used. TTS outperforms existing methods.}
\label{table:rot_ic13}
\vspace{-5.mm} 
\end{table}

\subsection{Weakly-Supervised Results}

In Table \ref{table:box} we show results using different supervision types.
Unlike most of the previous methods, \TTSb, \TTSs~ and \TTSw~ output axis-aligned bounding boxes and not polygons, and are therefore evaluated by matching the bounding boxes of the ground truth polygons with our method's bounding boxes output, using a matching threshold for the axis-aligned IOU of 0.5.

We show that this change has only a minor affect on the evaluation by demonstrating on a previous method (MTS-V3 \cite{spotterV3}). Using the published model, we compute bounding boxes for the polygonal outputs and evaluate with our axis-aligned evaluation (Table \ref{table:box}). We compare the results of the axis-aligned evaluation to the method's official polygonal evaluation results (Table \ref{table:poly}).
The axis-aligned evaluation slightly lowers the results in comparison with the polygonal evaluation, so the results in Table \ref{table:box} can be compared to the polygonal evaluation results in Table \ref{table:poly}. 
Training using only the synthetic data (\TTSs) results in very low performance. In comparison, using our weakly-supervised training scheme (\TTSw) improves the results significantly, reaching competitive results to fully-supervised state-of-the-art methods while only using the transcription supervision on the real datasets.
Using bounding box supervision (\TTSb) improves results even further and reaches state-of-the-art results. When directly comparing \TTSb~ and \TTSp~ (Table \ref{table:poly}), we see that the polygonal annotations do not improve the results, and sometimes even degrade them. This is due to the fact that the model is trained without polygons, and the segmentation head is trained afterwards, with the rest of the model weights frozen. 
We believe further optimization of this head can improve the results, but this is not the focus of this work.

\subsection{Robustness to Rotation and Curvature}
\label{sec:Robustness}

We evaluate TextTranSpotter's robustness to rotation by testing it on the Rotated ICDAR 2013 dataset. Table \ref{table:rot_ic13} shows our model improves performance on this dataset compared to previous approaches, even though our models are trained using only bounding boxes (the segmentation head is trained separately, as described in Sec. \ref{sec:arch}). Using bounding boxes causes more background text and noise to enter the recognition head, and there is no explicit information about the orientation of the text. However, since TextTranSpotter uses the transformer output as an input to both recognition and detection heads, it is able to ignore the irrelevant information and produce the correct transcript.
Fig. \ref{fig:results} shows \TTSw and \TTSp~ performance on challenging examples. Our method is able to handle large variations in rotation and font scale as well as cases with large overlaps between the bounding boxes.

\begin{table}[]
\centering
\begin{tabular}{@{}l|ccc@{}}
\toprule
\multirow{2}{*}{TTS Architecture} & \multicolumn{3}{c}{Detection} \\ \cmidrule(l){2-4} 
                        & P        & R        & F       \\ \midrule
EncDec. + det.       & 88.4     & 82.8     & 85.5    \\
\textbf{EncDec. + det. + recog.}                     & \textbf{90.9}     & \textbf{84.4}     & \textbf{87.6}    \\ \bottomrule
\end{tabular}
\vspace{-1.5mm}
   \caption{\textbf{Detection ablation.} Detection precision, recall, and F-measure of TTS, with and without the recognition head, on Total-Text dataset. It can be seen that optimizing the query embedding for the recognition task, improves the detection task.}
\label{table:det-ablation}
\vspace{-1.mm}
\end{table} 
\begin{table}[]
\centering
\begin{tabular}{@{}l|l|lc@{}}
\toprule
\multirow{2}{*}{Matching Criter.} & \multirow{2}{*}{Recog. Head (params)} & \multicolumn{2}{c}{End-to-End} \\ \cmidrule(l){3-4} 
                                  &                                & None       & Full       \\ \midrule
det. + cls.                         & linear (3.4M)                  & 73.6       & 83.6       \\
det. + cls.                         & RNN (2.8M)                     & 74.0       & 84.5       \\
\textbf{det. + cls. + recog.}                 & \textbf{RNN (2.8M)}                     & \textbf{75.8}       & \textbf{84.5}       \\ \bottomrule
\end{tabular}
\vspace{-1.5mm}
   \caption{\textbf{Recognition head and matching ablation.} End-to-end results of models on the Total-Text dataset, trained in a fully-supervised manner, using linear and RNN recognition heads, and with and without the recognition criterion. It can be seen that using the recognition matching criterion improves performance, and that the RNN recognition head is preferable to the linear head.}
\label{table:rec-ablation}
\vspace{-2.mm}
\end{table}

\subsection{Annotation Cost Study}
\label{sec:annotations}
To estimate the annotation time required for each labeling method, we conducted a user study on 100 images out of the TotalText dataset \cite{totaltext} with 9 annotators. Each user was asked to annotate different images with polygons and transcriptions, bounding boxes and transcriptions, or only transcriptions. The results as presented in Fig. \ref{fig:anno-time} show that the average annotation time per instance is 14.3, 10.6 and 4.6 seconds for polygon, bounding box and transcription only annotations respectively. This is consistent with the results by Zhao \etal \cite{weak_detection} which show that the average annotation time per image on the ICDAR-ArT dataset \cite{icdar2019} is 60 and 39 seconds using polygons and bounding boxes respectively (without transcriptions).

\subsection{Ablation Study}
\label{sec:ablation}
We test the detection performance of the original Deformable DETR \cite{defdetr} compared to the fully-supervised TextTranSpotter, shown in Table~\ref{table:det-ablation}. We train both models in the same manner, with the significant differences being the Text Hungarian loss for TTS and the recognition head. The models are evaluated on the Total-Text dataset using the standard text detection metrics. \TTSb~ outperforms the vanilla Deformable DETR model, improving both recall and precision of the model. This experiment highlights the benefit of mutually optimizing the detection and recognition tasks, in comparison to training separate standalone models for each task.

Next, we study the impact of the Text Hungarian Loss proposed in Sec. \ref{sec:hungarian}. We train our fully supervised model using two different matching criteria for the Hungarian matching algorithm; detection and classification, as presented in DETR \cite{detr}, versus detection, classification and recognition, as in our Text Hungarian Loss (Sec. \ref{sec:hungarian}). We evaluate the models for both end-to-end and detection on Total-Text, and show our results in Table \ref{table:rec-ablation}. The text matching criterion improves results, mainly for recognition. Therefore, using it improves performance for the end-to-end setting without a lexicon. When using a lexicon, the improvement to the recognition performance is less significant and the end-to-end results remain the same. We use the full matching criteria for our fully-supervised training.

The query embeddings which go into the recognition head in TTS are one dimensional and have no spatial or sequential structure, in contrast to previous recognition architectures. In addition, the recognition head is trained without using the ground truth transcription during the forward pass like previous approaches, since the matching is performed only at the end of the forward pass. 
Taking into account these two significant changes, we aim to study the contribution of using an RNN compared to using linear layers. The results using the two different recognition heads are presented in Table \ref{table:rec-ablation}. Using a linear head lowers the results compared to an RNN head, showing that the recurrent output formulation is beneficial for the recognition task while reducing the number of parameters in the recognition head. 
\section{Conclusions}
\vspace{-1.5mm}
\label{sec:conclusions}
We presented the first text spotting framework that can be trained in both fully- and weakly-supervised settings. By using a transformer encoder-decoder to learn a joint representation for the recognition and detection tasks, we can forgo much of the expensive annotations that are required in other approaches, and trade-off model accuracy vs. annotation time. The transformer's attention mechanism helps achieve accurate results on difficult cases, such as curved, rotated, dense, and even upside-down text. Our novel Text Hungarian Loss includes the recognition information in the detection optimization and permits training without the detection supervision altogether. Our method achieves state-of-the-art results on several benchmarks in the fully-supervised approach, and competitive results in the weakly-supervised setting. 
We hope that this work will open the door to new research directions in the field of text spotting, and to new views regarding which annotations are truly required for this task, examining trade-offs and combinations of weakly and fully supervised data.

%%%%%%%%% REFERENCES
{\small
\bibliographystyle{ieee_fullname}
\bibliography{egbib}
}

\end{document}